\documentclass[preprint,12pt, times]{elsarticle}

\usepackage{graphicx}%
\usepackage{multirow}%
\usepackage{amsmath,amssymb,amsfonts}%
\usepackage{amsthm}%
\usepackage{mathrsfs}%
\usepackage[title]{appendix}%
\usepackage{xcolor}%
\usepackage{textcomp}%
\usepackage{manyfoot}%
\usepackage{booktabs}%
\usepackage{algorithm}%
\usepackage{algorithmicx}%
\usepackage{algpseudocode}%
\usepackage{listings}%
\usepackage[normalem]{ulem}%
\usepackage{booktabs}%
\usepackage{longtable}
\usepackage{subcaption}
\usepackage{adjustbox}
\usepackage{array}
\usepackage{url}

\newcommand{\quotes}[1]{``#1''}

\begin{document}

\begin{frontmatter}

\title{Dynamic Bayesian Networks for Predicting Cryptocurrency Price Directions: Uncovering Causal Relationships}


\author[label0]{Rasoul Amirzadeh} 

\affiliation[label0]{organization={School of Information Technology, Deakin University},
            addressline={Waurn Ponds Campus}, 
            city={Geelong},
            postcode={3216}, 
            state={Victoria},
            country={Australia}}

\author[label0]{Dhananjay Thiruvady} 
\author[label0]{Asef Nazari} 

\author[label1]{Mong Shan Ee} 
\affiliation[label1]{organization={Deakin Business School, Deakin University}, 
            addressline={Burwood}, 
            city={Melbourne},
            postcode={3125}, 
            state={Victoria},
            country={Australia}}

\begin{abstract}
Cryptocurrencies have gained popularity across various sectors, especially in finance and investment. Despite their growing popularity, cryptocurrencies can be a high-risk investment due to their price volatility. The inherent volatility in cryptocurrency prices, coupled with the effects of external global economic factors, makes predicting their price movements challenging. To address this challenge, we propose a dynamic Bayesian network (DBN)-based approach to uncover potential causal relationships among various features including social media data, traditional financial market factors, and technical indicators. Six popular cryptocurrencies, Bitcoin, Binance Coin, Ethereum, Litecoin, Ripple, and Tether are studied in this work. The proposed model's performance is compared to five baseline models of auto-regressive integrated moving average, support vector regression, long short-term memory, random forests, and support vector machines. The results show that while DBN performance varies across cryptocurrencies, with some cryptocurrencies exhibiting higher predictive accuracy than others, the DBN significantly outperforms the baseline models.

\end{abstract}


\end{frontmatter}
\section{Introduction}\label{sec1}

The cryptocurrency market has emerged as an important player in global financial markets despite its relatively short lifespan~\cite{maasoumi2021contrasting}. Since the inception of cryptocurrencies in 2012,\footnote{Initiated by the \quotes{Bitcoin: A Peer-to-Peer Electronic Cash System} paper by Nakamoto Satoshi~\cite{nakamoto2008peer}.} the market capitalisation has soared, increasing from 17 billion in early 2017 to 2.6 trillion US dollars in March 2024.\footnote{coinmarketcap.com} This impressive growth underscores the significance of cryptocurrencies and their impact on attracting a large number of investors.

Despite the attractiveness of the cryptocurrency market, several challenges affect the profitability of cryptocurrency trading~\cite{li2023profitability}. Cryptocurrencies face common challenges encountered by traditional financial assets, such as accurate price prediction and global economic and non-economic influences~\cite{charfeddine2020investigating}. These challenges are primarily due to the inherently complex, nonlinear, and noisy attributes of financial data~\cite{wang2021stock} and influences of factors such as market sentiment, regulatory changes, and macroeconomic indicators~\cite{ anamika2023does, nabilou2019central}. Additionally, cryptocurrencies encounter unique challenges due to their distinct features, which are not present in other financial markets, including mining difficulty, wallet and exchange security, blockchain energy consumption, and lack of international acceptance and regulation~\cite{sabry2020cryptocurrencies, valencia2019price,amirzadeh2024dynamic}. These factors contribute to higher volatility and add complexity to predicting cryptocurrency price movements~\cite{el2023stochastic}. Considering this complexity, providing investors with tools that can explain the relationships between these factors and predict price direction is influential. 
By simplifying the problem to classifying price movement direction, these tools can identify trends, improve predictive accuracy for financial assets~\cite{wang2015stock, quesada2022piecewise}, and generate buy and sell signals for trading~\cite{ismail2020predicting}.

Artificial intelligence (AI), particularly machine learning (ML), has proven effective in prediction across various fields~\cite{shaygan2022traffic}. Researchers and FinTechs\footnote{The term ``Fintech'' refers to individuals or companies that bring innovation and disruption to the financial industry by merging technological and financial capabilities~\cite{altan20future}.} are increasingly utilising ML techniques, such as long short-term memory~\cite{swathi2022optimal} and artificial neural networks~\cite{liu2022quantum}, to address the challenge of price prediction in the cryptocurrency market. However, finding an appropriate ML technique to achieve high accuracy is complex~\cite{amirzadeh2022applying}. In particular, the accuracy of prediction is sensitive to the choice of the model, sample size and corresponding hyperparameters~\cite{cummings2021subjectivity, jin2021modeling}. Additionally, many ML models are limited by their ``black-box'' nature, which lacks explainability and does not provide insights into the importance of features or the relationships between variables~\cite{aldrees2024evaluation}. These limitations are particularly challenging in the context of financial markets, where understanding the underlying factors driving predictions is crucial for making informed investment decisions.

To address these challenges, we use DBNs to predict the price direction of cryptocurrencies for several reasons. They offer a structured approach to modelling complex systems that change over time. They efficiently describe Markov processes, requiring only the current state to determine the probability of the next state, making them suitable for the volatile and complex cryptocurrency market~\cite{heine2020towards}.  Compared to some ML models used for classification, which are limited to predicting only the direction of market movement without accounting for the magnitude of price movement directions~\cite{valencia2019price}, DBNs offer the advantage of providing a probabilistic inference for price movement directions, estimating the magnitude and likelihood of such movements. 
One of the key advantages of DBNs is their explainability, which facilitates understanding the relationships among various factors. Specifically, DBNs can identify the most influential features by establishing causal relationship networks among them and offering a graphical model to represent the dynamics of systems~\cite{koutsoukos2009distributed}. This approach offers deeper insights into the drivers of cryptocurrency price and their impact on price fluctuations~\cite{zhang2016multiobjective}. Despite all these important attributes, DBNs are rarely used in the domain of cryptocurrency.

We employ DBNs to predict the direction of price movements in cryptocurrencies. In addition to Bitcoin, the market leader, the study investigates the effectiveness of DBNs in predicting the daily price directions of five popular altcoins, which are mainly chosen considering market capitalisation and data availability.
The performance of the DBN models is compared with five frequently used baseline models, namely autoregressive integrated moving average (ARIMA), support vector regression (SVR), random forests~(RF), long short-term memory~(LSTM), and support vector machines~(SVM) models. Moreover, the research explores various features to determine if increasing the number of features fed into the DBN leads to a monotonic improvement in prediction accuracy. By examining how different feature categories influence the performance of DBNs, we aim to enhance the accuracy of market direction predictions. The contributions of our research are:

\begin{itemize} 
\item Analysing daily price direction movements for six popular cryptocurrencies—Bitcoin, Binance Coin, Ethereum, Litecoin, Ripple, and Tether—using AI methods to evaluate predictive accuracy and provide insights into market behaviour.
\item Investigating the impact of four feature groups—price information, social media activity, macro-financial data, and technical indicators—on the accuracy of cryptocurrency price predictions, highlighting their interplay and collective influence on predictive performance.
\item Proposing a DBN-based approach for predicting cryptocurrency prices by analysing causal relationships among market factors. This methodology generates probabilistic ``buy" and ``sell" signals, helping investors make informed trading decisions.
\item Delivering explainable Bayesian networks that identify the most influential factors affecting price changes through causal inferential analysis, thereby alleviating the ``black box" problem faced by many AI models. 
\end{itemize}
    
The remainder of this paper is structured as follows. In Section~\ref{background}, a literature review is provided on the studies related to predicting cryptocurrency prices. Section~\ref{method} introduces DBNs. Section~\ref{proposed-system} outlines the selection of features.
Section~\ref{experimental} provides information on experimental design and data analysis procedures.
The findings of our study are discussed in Section~\ref{Results}. Finally, Section~\ref{Conclusions} provides concluding remarks on the study and suggests directions for future research.

\section{Related work \label{background}}
In this section, we review a set of recent academic publications concerning using ML models in predicting price movement directions. We first briefly consider the feature engineering side of devising ML models including incorporating technical indicators, social media data, and their impact on the accuracy of the models. We then survey the applications of Bayesian networks~(BNs) in predicting upward and downward trends in financial markets. 

Technical indicators are valuable tools for making predictions and trading decisions, offering financial market participants valuable insights into market trends. They are commonly used features in ML studies for predicting various financial markets, as seen in \cite{borovkova2019ensemble} and \cite{choudhry2008hybrid}. However, they have received less attention as input features in the cryptocurrency literature. For instance, \citet{huang2019predicting} build a tree-based model to assess Bitcoin return predictability, using 124 indicators categorised into overlap study, momentum, cycle, volatility, and pattern recognition. Their model shows predictive power for Bitcoin's narrow intraday ranges and outperforms the buy-and-hold strategy, demonstrating the utility of technical analysis in Bitcoin despite its value being driven mainly by non-fundamental factors.
Another study by \citet{alonso2020convolution} predicts trends for six cryptocurrencies—Bitcoin, Dash, Ethereum, Litecoin, Monero, and Ripple—using eighteen technical indicators calculated from one year of data. They classify one-minute trends into increase, neutral, or decrease categories with four neural network architectures, including convolutional neural networks and multilayer perceptrons. The study demonstrates that all cryptocurrencies can be predicted to some extent with technical indicators, with better performance for Bitcoin, Ethereum, and Litecoin. However, short-term predictions are limited by response times and liquidity issues.
In another study, \citet{akyildirim2021prediction} analysed the predictability of 12 cryptocurrencies using four ML algorithms, including support vector machine (SVM) and logistic regression. For their investigations, they use past price information and eight technical indicators, including the five-day relative strength index (RSI) and simple moving average (SMA) as features for their models. The results show that all four algorithms have an average classification accuracy consistently above the 50\% threshold for all cryptocurrencies.
 
The influence of popular social platforms, such as Twitter,\footnote{The company's name has changed to X, but we use the name that was in effect at the time of the study.}  on cryptocurrency price movements, has received substantial attention~\cite{ye2022stacking, zou2022multimodal} in the literature. In a work by~\citet{abraham2018cryptocurrency}, the authors predict changes in Bitcoin and Ethereum prices using Twitter and Google Trends data. According to their study, the volume of tweets is a predictor of price direction rather than sentiment. Moreover, they employ a sentiment analysis tool in the study and observe that the sentiment of tweets tends to remain positive, regardless of price changes in their investigation.
In another study, \citet{valencia2019price} explores price movement prediction for four cryptocurrencies—Bitcoin, Ethereum, Ripple, and Litecoin—using multilayer perceptron, SVM, and random forest models. They incorporate two feature groups: raw price data and Twitter sentiment data. The Twitter data is categorised into positive, negative, neutral, and compound sentiments. The study compares the models' performance based on independent and combined feature groups. Results indicate that Twitter sentiment data alone can predict cryptocurrency movements, with the highest accuracy achieved for Bitcoin. However, there is no universally accepted performance criterion, and different models show varying effectiveness across cryptocurrencies based on accuracy and precision.

Despite the considerable capacity of DBNs in modelling complex stochastic situations and detecting features with causal relationships, it has rarely been applied in analysing the cryptocurrency market. However, studies exploring BNs and DBNs in other domains highlight their versatility and effectiveness in various applications.
For instance, \citet{wang2015stock} use DBNs with nine macroeconomic factors to predict stock market trends in the US and China. Their method effectively detects market trend changes before actual turning points, however, it does not determine the direction of these trends, focusing instead on the timing of market shifts.
\citet{jangmin2004stock} apply a variant of DBNs to model price trends for 20 companies in the Korean stock market. Although the model does not surpass the buy-and-hold strategy, which performs well during a bull market, it does achieve better cumulative profit compared to the triple exponential average indicator. 
In another study, \citet{wang2017stock} explore the use of DBNs for predicting stock prices and generating profits. They analyse NASDAQ and the Stock Exchange of Thailand using five years of historical intraday closing-price data. Their DBN model demonstrates significant profit generation, statistically outperforming several benchmark strategies, including the buy-and-hold approach.

To predict stock index price movements, \citet{zuo2012up} employ BNs and compare their performance against the psychological line and trend estimation algorithms. The psychological line algorithm uses historical price data to forecast future price directions. The study finds that the BN model achieves an accuracy rate of around 60\%, which is approximately 10\% higher than the other investment strategies considered. Additionally, BNs demonstrate significantly greater profitability than the baseline models, highlighting their effectiveness in predicting price movements. In addition, \citet{malagrino2018forecasting} use BNs to assess the influence of foreign markets on Brazil's iBOVESPA index. They test two BN models with 24-hour and 48-hour timeframes to predict the next day's index direction. The 24-hour BNs achieve a mean accuracy of about 71\%, outperforming the 48-hour BNs, which have a 68\% accuracy rate. The study also finds that including fewer indices improves BN performance.

In summary, based on the reviewed studies, while technical indicators and social media data have shown promising results, there is a notable lack of comprehensive research on their application across various cryptocurrencies. Furthermore, the potential of DBNs for cryptocurrency prediction, along with their ability to provide explainable influence networks, remains underexplored, highlighting the need for further research.

\section{Dynamic Bayesian networks \label{method}}
BNs are a powerful tool for modelling complex systems and can be used to extract the underlying causal relationships between variables \cite{heckerman2008tutorial}.
They are a type of probabilistic graphical model (PGM) used to represent stochastic relationships between random variables through directed acyclic graphs (DAGs)~\cite{alameddine2011evaluation}. They are based on the Bayesian theory of conditional probability and can be used for reasoning, prediction, and decision-making in a wide range of fields from environmental management~\cite{death2015good}, maritime accidents~\cite{kuzmanic2021weight} to wind energy industry~\cite{adedipe2020bayesian} and credit assessment~\cite{masmoudi2019credit}. In a BN, nodes represent variables, and directed edges represent the conditional dependencies between variables. Each node in the network is associated with a probability distribution that describes the probabilities of possible values of that variable given the probabilities of values of its parent variables. These joint probability distributions over the nodes are represented by conditional probability tables (CPTs)
for each node~\cite {rohmer2020uncertainties}. BNs can be utilised to perform inference, which involves using the network structure of the variables in the system to make predictions or draw conclusions about the state of the system given some evidence or observations.  In particular, the capacity of a visual representation of dependencies in a system using DAGs makes BNs a versatile tool for communicating some properties of the system. They can handle missing data and hidden variables, and on top of that training of BNs inherently allows for avoiding overfitting \cite{heckerman2008tutorial}, which is a common problem in ML modelling.

\begin{figure*}[h]
  \centering
    \begin{subfigure}{.65\textwidth}
        \centering
\includegraphics[width=0.9\linewidth,height=0.9\textheight,keepaspectratio]{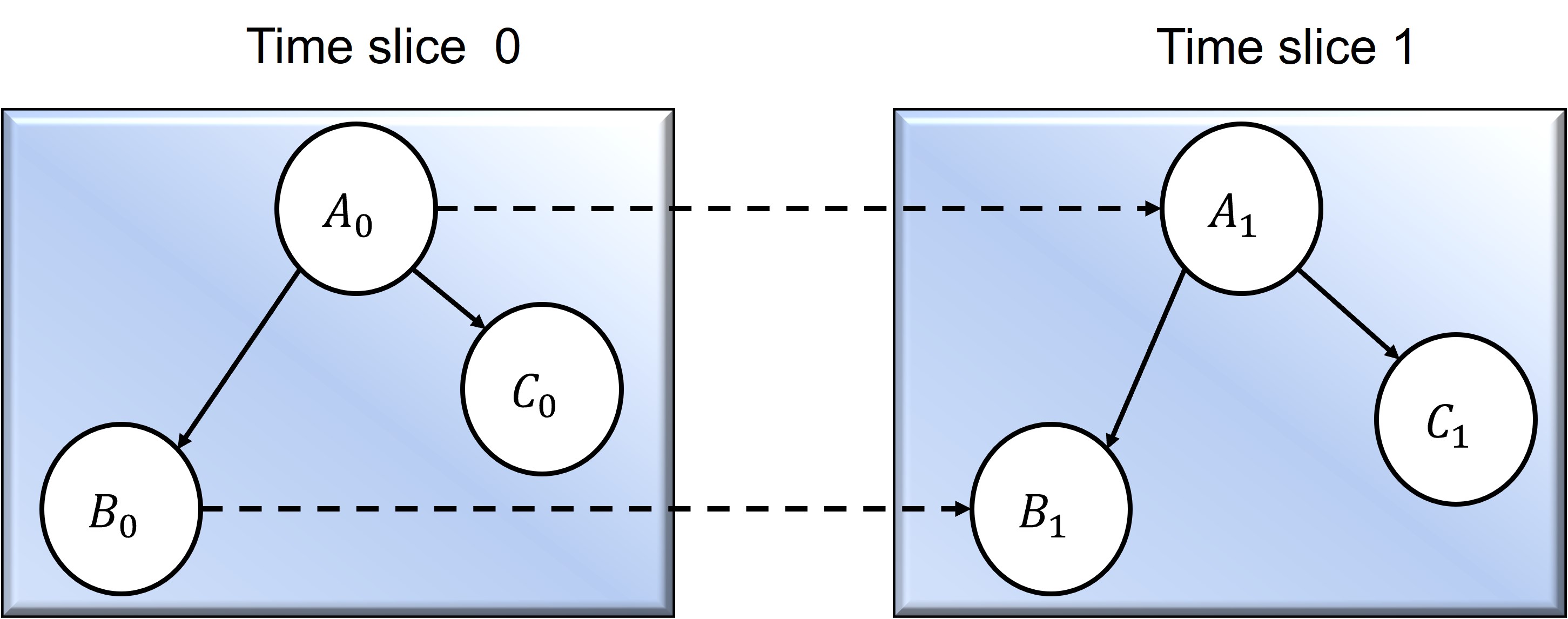}
        \caption{Unrolled 2TBN}
        \label{fig:dbn}
\end{subfigure}%
  \begin{subfigure}{.35\textwidth}
        \centering
\includegraphics[width=0.9\linewidth,height=0.9\textheight,keepaspectratio]{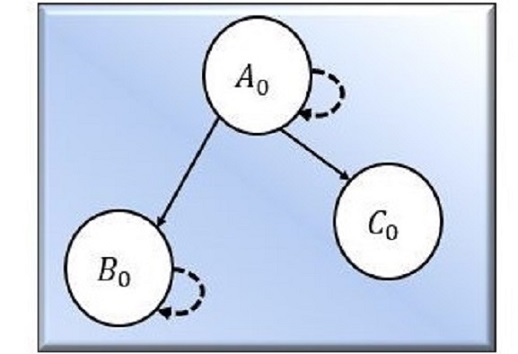}
        \caption{2TBN}
        \label{fig:dbnone}
  \end{subfigure}
  \centering
  \caption{An illustration of a DBN composed of two-time slices. Note that these two models represent the same DBN, where Sub-figure~(b) represents a compact version of Sub-figure~(a). While connections between $A_{0}$, $B_{0}$, and $C_{0}$ inside time slices $0$ represent intra-slice arcs (solid links), links between $A_{0}$ and $A_{1}$ and $B_{0}$ and $B_{1}$ from time slice $0$ to time slice $1$ are inter-slice arcs (dotted lines).
  }
\end{figure*}

DBNs are a type of BNs that can be used to model systems that change over time. Unlike static BNs, which only model dependencies between random variables at a single point in time, DBNs represent the temporal evolution of a system by establishing time-dependent dependencies between variables. In a DBN, each node represents a random variable at a specific point in time, and directed edges model the conditional dependencies between variables at the same or different time intervals \cite{shiguihara2021dynamic}. A DBN is composed of two parts: time slices and inter-slice arcs. A time slice represents the states of the system at a given time $t$ and is essentially an identical BN at each time step. Within a time slice, the relationships between variables are represented by intra-slice arcs. The second component of a DBN is inter-slice arcs, also known as temporal arcs. These arcs represent the relationships between variables within a time slice or between certain variables across time slices. These components are illustrated in Figure \ref{fig:dbn}. In addition, a basic assumption in constructing DBN is that the system follows a first-order Markov process, in which the state of the system at time $t$ only depends on its state at the previous time slice $t-1$. In other words, future temporal nodes at time $t$ are only connected to corresponding nodes at the previous time slice $t-1$ \cite{gao2014approximate,wu2015dynamic}.

The process of building a DBN is iterative. Each time slice requires the same structural form as the previous or next slice, and time slices reflect the change in probabilities of the variables \cite{zhang2023dynamic}. In particular, converting a BN to a DBN involves three main steps. The first step is to modify the BN structure to incorporate the dynamics of the process. Next, one needs to introduce a time parameter in the definition of the states of all nodes to describe the temporal relationship. Finally, the static BN is repeated for $n$ time steps and the belief in the system is updated for the given time step \cite{voronenko2020dynamic}.

From a mathematical perspective, a DBN is a pair ($\mathit{B_{0}}$,$\mathit{B_{\rightarrow}}$), where $\mathit{B_{0}}$ is a BN model that defines the prior network, and $\mathit{B_{\rightarrow}}$ is a two-time slice temporal BN (2TBN) which defines the relationship between two consecutive time slices through a transition probability table \cite{zhang2023dynamic}. The joint probability distribution of a DBN can be demonstrated as
$$P(X_{1:T}^{1:N}) = \prod_{t=1}^{T}\prod_{i=1}^{N}P({X_{t}}^{i}|Pa({X_{t}}^{i})),$$
\noindent where  $X_{t}^{i}$ is the $i^{th}$ node at time step $t$, and $Pa({X_{t}}^{i})$ represents the parent nodes of $X_{t}^{i}$ in the corresponding DAG. Furthermore, the conditional probability $P({X_{t}}^{i}|Pa({X_{t}}^{i})$ indicates that the transition probabilities are a product of the CPTs in the 2TBN, where $T$ is the full-time horizon ($T = 5$ in this study), and $N$ in the number of nodes in $X_{t}^{i}$. Once the probabilities on nodes in a DBN are determined through the joint probability distribution calculation, different forms of reasoning and inferencing such as prediction, diagnosis, or decision-making can be performed.


\section{Development of a DBN-based model for cryptocurrency price prediction}\label{proposed-system}

This section describes  the process for developing the DBN model for cryptocurrency price movement directions. It briefly discusses the rationale for selecting particular features and then provides a detailed explanation of the process for developing DBNs based on these features.

\begin{figure*}[h]
  \centering
\includegraphics[width=0.86\linewidth,height=0.85\textheight,keepaspectratio] {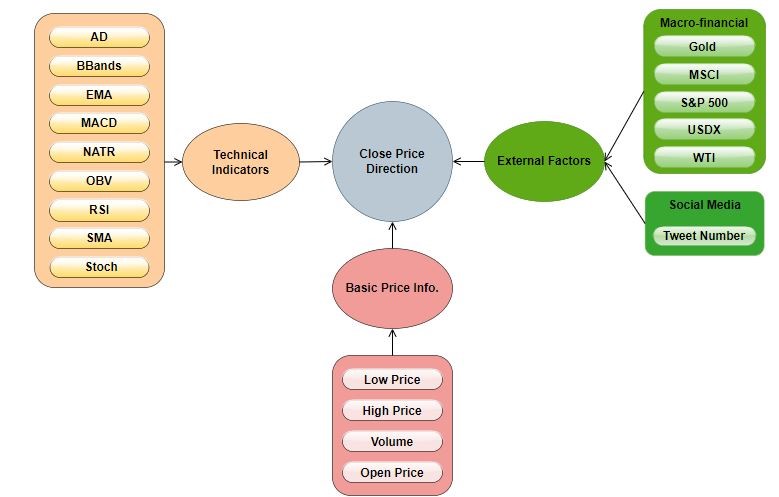}  \caption{ 
This conceptual framework depicts the feature engineering approach used in the study, outlining three distinct feature categories that impact the direction of closing price movements, depicted in a grey circle. Basic price information is shown in pink, encompassing open, high, low and volume data. The second group, coloured in amber, consists of nine specific technical indicators. The third category, external factors, illustrated in green, comprises macro-financial factors, including five financial assets and social media impact measured by daily tweet volume related to each cryptocurrency
}
  \label{fig:con_model}
\end{figure*}

To predict cryptocurrency price directions, we analyse several feature categories to assess their impact on the model's accuracy. Figure~\ref{fig:con_model}  presents the conceptual framework of our proposed cryptocurrency price prediction model, illustrating the structure of the selected feature groups. Our framework incorporates three key types of features:

\begin{itemize}

    \item \textbf{Basic Price Data}: This includes daily open, high, low, and close prices and trading volume~(OHLCV). This group of features provides a detailed snapshot of a financial asset's status on a given day ~\cite{motard2022hierarchical}.

    \item \textbf{Technical Indicators}: We use various technical indicators derived from price data through mathematical formulas~\cite{manujakshi2022hybrid}. Technical indicators help identify market trends and potential buy or sell signals, offering insights into the strength of movements, trend reversals, and price volatility, thereby enhancing our model's predictive power. Additional details and applications of these indicators can be found in \cite{ alonso2020convolution,srivastava2021deep}. 

    \item \textbf{External Factors}: Cryptocurrency prices are also influenced by external factors, which can be categorised into two main groups: adoption and attractiveness, and macro-financial drivers \cite{ciaian2016economics, poyser2019exploring}. Adoption and attractiveness, rooted in behavioural finance,\footnote{Behavioural finance is a relatively new field that explores psychological reasons behind investor decision-making~\cite{konigstorfer2020applications}.} pertain to the market’s investment appeal and investor intentions. Macro-financial drivers encompass various traditional financial assets and economic indicators, such as currencies, commodities, and stock indices, which are commonly studied in relation to cryptocurrency markets \cite{corbet2018exploring, charfeddine2020investigating}. 
    A short summary of all features used in this study is provided in~\ref{feature_definitions}.
\end{itemize}

In this study, we first build BNs to identify and understand causal relationships within the training data. These BNs help us capture how different price features interact and generate probability distributions for different states of the system. However, BNs only show relationships at a single point in time and they do not capture the temporal evolution of these factors. 

\begin{figure*}[t]
  \centering
\includegraphics[width=0.6\linewidth,height=0.6\textheight,keepaspectratio]{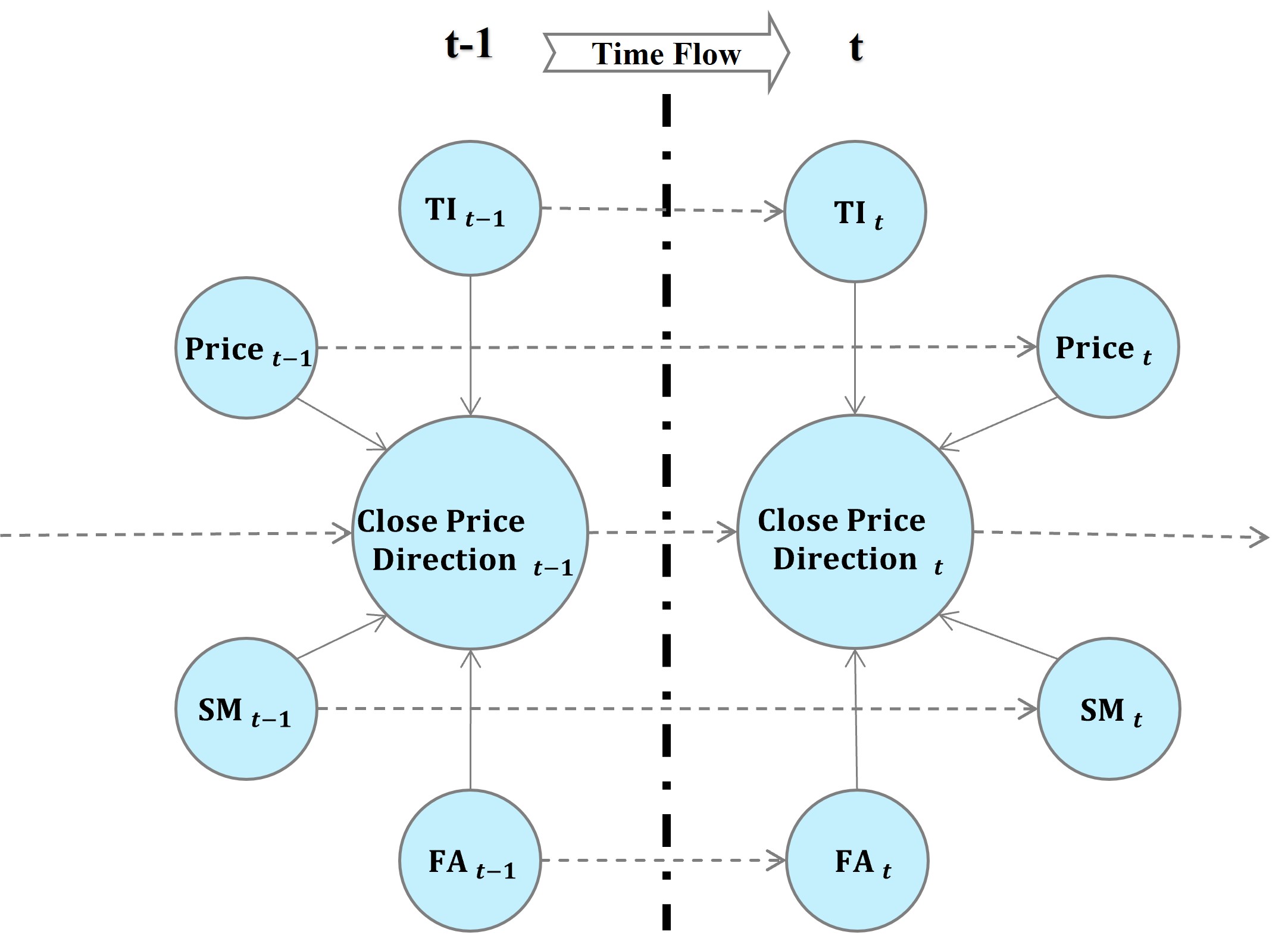}
  \caption{An illustration of the proposed DBN model at two consecutive time steps for price direction prediction based on the dynamic interactions between corresponding feature groups: abbreviated as TI for technical indicators, Price for OHLV features, SM for social media (tweet numbers), and FA for financial assets. The model includes four feature groups at each of the two consecutive time steps, and inter-slice arcs (dotted lines) connect corresponding nodes between the time steps. Note that the intra-slice arcs (solid lines) connected to close price direction are only for illustrative purposes and do not indicate actual causal relationships among the feature groups in our constructed DBNs. Potentially there could be intra-slice arcs between groups of features}
  \label{fig:db}
\end{figure*}

To address this limitation, we convert  BNs into DBNs. Unlike BNs, DBNs model the system over multiple timeframes, enabling us to track how probabilities evolve over time. Specifically, our DBNs use a time window of the last five days of price data to predict today’s price direction. The architecture of our proposed DBN is shown in Figure~\ref{fig:db}. In this model, we assume that each feature group is independent of one another, meaning no inter-slice arcs between two consecutive timeframes. This simplifies the model by focusing on temporal dependencies within individual timeframes, while still capturing causal relationships within each timeframe. This approach helps to balance model complexity and interpretability.

\section{Experimental design\label{experimental}}
The study aims to predict the daily price direction of six cryptocurrencies using DBNs. Additionally, it explores the impact of various combinations of feature groups on the model's price prediction performance.
To achieve this goal, four distinct feature groups are created as follows:
\begin{itemize}
    \item The first combination includes only the OHLCV data.
    \item   The second combination incorporates OHLCV with external price factors containing Twitter data and traditional financial assets.
    \item  The third combination uses OHLCV data along with nine technical indicators.
    \item  The fourth combination includes all features.
\end{itemize}
Traditional financial asset data is generally unavailable on weekends, unlike cryptocurrencies and Twitter data. Therefore, weekday data points from these three sources are aligned to create a unified database for each cryptocurrency
enabling numerical analysis. Table~\ref{tab:group} outlines the different combinations of features with each combination assigned a unique group number, ranging from 1 (with the fewest number of features) to 4 (with the largest number of features). After creating the four groups of features, they were preprocessed using the min-max normalisation method to balance the input data ranges. Min-max normalisation has been reported to deliver satisfactory performance in supervised and unsupervised learning tasks \cite{wijaya2021dwtlstm, shantal2023novel}. 
\begin{table*}[h]
\centering
\caption{The list of feature group combinations and the number of individual features in each combination 
}
\resizebox{\columnwidth}{!}{%
\renewcommand{\arraystretch}{1.2}
 \begin{tabular}{clcl}
\hline
\textbf{No} & \multicolumn{1}{c}{\textbf{Groups of features}} & \textbf{No. of features} & \multicolumn{1}{c}{\textbf{Abbr.}} \\ \hline
1 & OHLCV & 5 & OHLCV \\ \hline
2 & OHLCV and  external factors & 11 & OHLCV-EF \\ \hline
3 & OHLCV and technical indicators & 17 & OHLCV-TI \\ \hline
4 & OHLCV,  external factors, and technical indicators & 23 & OHLCV-EF-TI \\ \hline
\end{tabular}
\label{tab:group}
}
\end{table*}

Since the price data are continuous, and we predict future price directions, which are categorical variables, it is necessary to categorise the data prices into several market states. An appropriate choice of market states is dependent on the model choice or data and impacts the model's performance. 
For instance, some studies use three states for market trends, including down, steady, and up, while others may consider two additional trend states of strong down and strong up~\cite{zou2012influence, stefanovska2020effects}.  Based on the results presented in \cite{amirzadeh2023modelling},  the two states of ``Down'' and ``Up'' provide the best model performance for BNs applied on cryptocurrencies. Consequently, the data is labelled into these two states for market data, where each data record was labelled as ``Down'' if today's record value was lower than the previous day's, and ``Up'' otherwise. Equation~\ref{eq:trend} outlines the labelling approach for the data records. 
\begin{equation}
\label{eq:trend}
\text{Label for time step}(t+1)=
\left\{\begin{matrix}
 Down, &if & C_{t} < C_{t+1} \\ 
 ~Up, ~~~~ &if & C_{t} \geq C_{t+1}  \\
\end{matrix}\right.
\end{equation}
\noindent where $C_{t+1}$ means the data record of the next day, and $C_{t}$ is the data record of the current date. Furthermore, the labelled data was split into training and testing sets, with 67\% of the data allocated for training and 33\% for testing. This split ratio is widely adopted for data mapping and independent accuracy assessment~\cite{lyons2018comparison}.

To implement the DBNs, we utilise the GeNIe software package through the PySMILE wrapper. PySMILE is a Python-based package that allows for Bayesian inference and modification of Bayesian networks using Python~\cite{bayesfusion2017genie}. Therefore, we fed the training set of each feature group into our implemented PySMILE code to learn the structure and parameters of the corresponding DBNs. Once the DBNs were constructed from the training set, a five-day moving window was generated from the test set and fed as inputs to the DBNs to predict the directions of price. Specifically, the last five-day data were used as inputs to our DBNs to predict the closing price. 

We compare the prediction accuracy of our proposed DBNs models against five baseline models. ARIMA and SVR are well-known and frequently used models for accurately predicting time series in various nonlinear systems, including economic and financial systems~\cite{fawzy2020comparison}. Moreover, we use LSTM, RF, and SVM, which are among the most widely used methods for price prediction~\cite{htun2023survey}. To achieve optimal predictions from these models, we employed parameter optimisation methods, including GridSearchCV \cite{alhakeem2022prediction} and grid search logic~\cite{chivukula2020cryptocurrency}, to fine-tune their parameters. 
 Furthermore, since the outputs of certain models, such as ARIMA, are continuous values, we labelled the price prediction as ``Up'' or ``Down'' based on Equation~\ref{eq:trend}.

To evaluate the performance of the models, we compare the predictions generated by DBNs, ARIMA, SVR, LSTM, RF, SVM models to the actual price directions using the precision metric.
The Precision metric is defined as a measure of exactness or accuracy, and it is computed as the ratio of true positives to all positive predictions (true and false positives)~\cite{liji2018improved}. True positives are instances correctly identified as positive by the model, whereas false positives are instances incorrectly identified as positive, actually being negative. Precision is chosen as the evaluation metric to ensure the models accurately identify the market direction, reducing the risk of incorrect price movement predictions and potential investment losses.
 Mathematically, the precision metric is defined as follows. 
 

$$\text{Precision} = \frac{TP}{TP+FP}$$
where TP is the number of true positives, and FP is the number of false
positives.

To ensure a consistent and standardised analysis of numerical findings, we use precision as a metric. This approach allows for a clear and uniform evaluation of model performance, simplifying comparisons and ensuring that all results are assessed using the same criteria. The precision metric measures the number of true positives when the model correctly predicts the positive records relative to the total number of positive predictions~\cite{quiroz2020image}.

\subsection{Data}

In this research, six cryptocurrencies are examined. Bitcoin was chosen as it is the most popular cryptocurrency and the altcoins chosen are ones that consistently ranked among the top ten cryptocurrencies by market capitalisation for several years. Collectively, the market capitalisation of five studied altcoins represents over 30\% of the cryptocurrency market at the beginning of 2023. Additionally, there are years of data available for these coins, which allows for a robust statistical analysis of the unpaired correlations between these cryptocurrencies and traditional financial assets. Specifically, we select coins that consist of 1,100 daily data records, equivalent to around four years worth of data from January 2018 to November 2023.

To obtain the daily price data for both cryptocurrencies and traditional financial assets, we use the yfinance Python package that connects to the Yahoo Finance website~(https://finance.yahoo.com/). Additionally, we extract the daily tweet count associated with each cryptocurrency from www.bitinfocharts.com.{\footnote{Both websites were accessed in May 2023.} The analysis of the impact of the social media data on Tether is excluded because this coin’s daily tweet data is unavailable. Descriptive statistics and plots for the price data of cryptocurrencies, traditional financial assets, and tweet numbers are provided in ~\ref{Descriptive} for further details.

To conduct the experiments, the data is divided into training and testing sets using a 67\% to 33\% ratio, which is a commonly used split ratio~\citep{lyons2018comparison}, followed by preprocessing of the data. Given the daily frequency, the number of missing entries is minimal, especially for the price data from Yahoo Finance. However, there are instances of null inputs in the Twitter data, which we remove before starting the analysis.


\section{Results and discussion}
\label{Results}

To assess the efficacy of our models, we compare the results of their prediction against those of baseline models, namely ARIMA, SVR, LSTM, RF, and SVM. We also investigate the impact of feature engineering on our DBNs’ performance. The experiments are conducted on a Windows 10 Enterprise operating system, running an Intel i7-Core(TM) CPU @ 1.90GHz, 2.11 GHz processor, and 16.0 GB of RAM.
We implement the baseline models in Anaconda and Python via the Sklearn, Statsmodels, and Keras libraries. They provide essential tools and functionalities for building, training, and evaluating various machine learning and statistical models.


Table~\ref{tab:preds} summarises the performance results for the best-performing DBN model for each coin among those evaluated, based on various features, as well as for five baseline models. These results pertain to predicting the next day's closing price of a cryptocurrency. It should be noted that the baseline models use only the close price time series data of cryptocurrencies. Additionally, the ‘Average for Model’ row
presents the average precision of each model for all cryptocurrencies.
\begin{table*}[h]
\caption{A comparison of daily price direction predictions using DBN, ARIMA, SVR, LSTM, RF, and SVM based on precision percentage. Precision is the measure of accuracy, calculated as the ratio of true positives to all positive predictions (true and false positives)
}
\label{tab:preds}
  \resizebox{\columnwidth}{!}{
\renewcommand{\arraystretch}{1.2}
\begin{tabular}{lcccccc}

\hline
& \multicolumn{6}{c}{\textbf{Model}} \\ \hline
& DBN & ARIMA & SVR & LSTM & RF &  SVM \\ \hline
Bitcoin  &  \textbf{72.19} & 48.90 &49.61& 44.96 &49.02&53.00\\ \hline
Binance Coin  &  \textbf{73.63} & 61.81 & 57.92& 43.01 &51.54&47.21\\ \hline
Ethereum  & \textbf{73.86}& 63.27 & 59.54& 44.00&47.83&48.73\\ \hline
Litecoin  &  \textbf{70.56} & 69.35 & 51.51& 44.33&52.35 &45.31\\ \hline
Ripple  & \textbf{77.00} & 48.64 & 46.35& 47.00&54.75 &51.31\\ \hline
Tether &\textbf{57.57} &39.81 & 41.54& 49.53 &53.31 & 52.69 \\ \hline
\textbf{Average~(sd)} &\textbf{70.81~(6.45)}  &55.30~(10.52) &51.08~(5.77)&45.47~(2.33) & 51.47~(2.95) & 49.71~(3.84)\\ \hline
\end{tabular}
}
\end{table*}

The best-performing model for each cryptocurrency is the DBN, outperforming all baselines despite their extensive use in the literature in predicting time series data.  Considering the baseline models, ARIMA and RF exhibit generally superior precision compared to the other models, with average scores of 55.30 and 51.47, respectively. Conversely, LSTM demonstrates low performance in predicting the daily price direction of cryptocurrencies.

Among all the coins, Ripple is the cryptocurrency with the highest overall prediction accuracy, considering precision,
across the models used with a maximum precision rate of 77\%. This observation is in line with the close
price plot of Ripple in Figure~\ref{fig:plots_price}, the details of which are provided in~\ref{Descriptive}, where the coin exhibits relatively low volatility across the period considered. On the other hand, predicting the price direction of Tether is more challenging, with a maximum
precision of 57.57\%.
Comparing Bitcoin's precision to altcoins for the DBN model, it is evident that certain altcoins~(Binance Coin, Ethereum, and Ripple) outperform Bitcoin in terms of prediction precision with DBN. This could suggest that DBNs can capture altcoin behaviours more effectively than Bitcoin's.
Figure~\ref{fig:percision} shows the best-performing DBNs based on the highest precision in Table~\ref{tab:preds}.\footnote{To further understand the DBN process,~\ref{sampleDBN} provides a detailed examination of DBNs for Ethereum and Tether, focusing on their node interactions and the impact of fixed states on the target node's probability changes.}

\begin{figure*}[h]
  \centering
\includegraphics[width=\linewidth,height=\textheight,keepaspectratio]{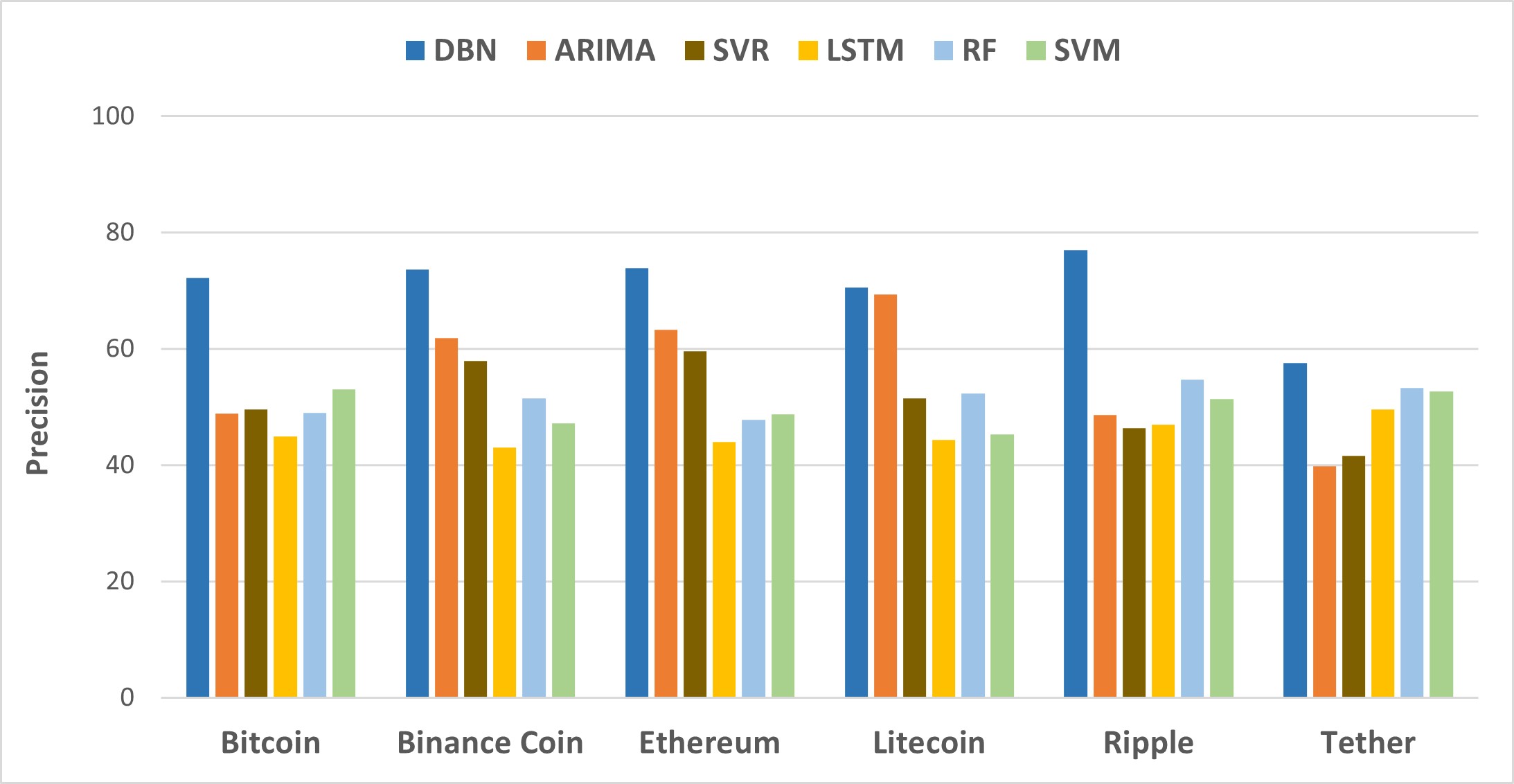}
 \caption{Precision of the proposed DBNs, ARIMA, SVR, LSTM, RF, and SVM for predicting price movement directions of cryptocurrencies}
\label{fig:percision}
\end{figure*}

\subsection{Evaluating the influence of the combination of features}

We investigate the impact of different combinations of features on the performance of DBNs. Table~\ref{tab:preds-DBNs} summarises the effect of feature engineering on the performance of the DBNs. Columns two to five correspond to the different feature sets used for building the DBNs. Each column represents a specific group of features, denoted by an abbreviation, defined in Table~\ref{tab:group}. 


 \begin{table*}[h]
\caption{ 
Overall comparison of daily market direction prediction using DBN models based on different groups of features considering the precision metric. OHLCV stands for Open, High, Low, Close, Volume; EF refers to external factors; TI denotes technical indicators. Each group of features is assigned a distinct group number
}

\label{tab:preds-DBNs}
  \resizebox{\columnwidth}{!}{
   \renewcommand{\arraystretch}{1.2}
\begin{tabular}{llllll}
\hline
 & \multicolumn{4}{c}{\textbf{DBN Feature Set Group}} \\ \hline
 & OHLCV  & OHLCV-EF &  OHLCV-TI &  OHLCV-EF-TI &  Best DBN  \\ \hline
Bitcoin  & 71.67 & 68.38 & \textbf{72.19} & 70.23 & DBN~(OHLCV-TI)\\ \hline
Binance Coin  & 59.27 & 71.02 & \textbf{73.63} & 69.98 &DBN~(OHLCV-TI)\\ \hline
Ethereum  & \textbf{73.86} & 68.02 & 71.83&71.83 &DBN~(OHLCV)\\ \hline
Litecoin  & 64.21 & \textbf{70.56} & 70.56 & 70.56  &DBN~(OHLCV-EF)\\ \hline
Ripple  &72.87 & 66.67 & \textbf{77.00} &72.87& DBN~(OHLCV-TI)\\ \hline
Tether & \textbf{57.57} & 57.57 & 55.09 & 50.12&DBN~(OHLCV) \\ \hline
\textbf{Average~(sd)} & 66.57~(6.57)&67.04~(4.49) &70.05~(6.99) &67.6~(7.88)&  \multicolumn{1}{c}{70.80~(6.23)}\\ \hline
\end{tabular}
}
\end{table*}

Table~\ref{tab:preds-DBNs} shows that there is no single DBN that performs best for all the coins considered in the study. This is consistent with other studies such as the study by \citet{valencia2019price},  indicating that different cryptocurrencies have varying specifications. This highlights the significance of feature engineering, as it allows for selecting and optimising specific feature groups tailored to each cryptocurrency, resulting in enhanced performance. 
Nonetheless, the best performance for Ethereum and Tether are obtained using only OHLCV data, whereas the DBN~(OHLCV-EF-TI) with the full set of features never outperforms the other DBNs. Similar results are observed for Binance Coin, Bitcoin and Ripple, where DBN~(OHLCV-EF-TI) does not exhibit superior performance compared to other DBN models. Thus, identifying the set of features for each coin must be done independently to achieve the best results. In addition, to compare the different versions of DBNs, the last row of Table~\ref{tab:preds-DBNs} shows the average precision of each model across all cryptocurrencies. Among the four different settings, the DBN with OHLCV and technical indicators has a slightly higher average precision.
The combination of these two feature sets is also selected three times as the best-performing model, for Binance coin, Bitcoin and Ripple. This underscores the potential importance of these features in enhancing model performance.



We see that on average, the combination of OHLCV and technical indicators yields the best results for all cryptocurerncies~(average precision $70.05\%$). Conversely, DBNs built solely on OHLCV achieve the lowest performance~(average precision $66.57\%$). However, it should be noted that adding more features to the DBN models does not always result in further improvements, but sometimes decreases the performance. For example, the DBN model constructed with all features~(OHLCV-EF-TI) for Binance Coin performs worse ($69.98\%$) than the DBN model that only contains OHLCV and technical indicators~(OHLCV-TI) with $73.63\%$ precision. Hence, including external factors alongside other groups of features may negatively impact the precision of the DBN model for Binance Coin.


The influence of technical indicators on DBN performance varies among cryptocurrencies. For Binance Coin, Litecoin, and Ripple, introducing technical indicators to the basic DBN model~(OHLCV) 
substantially substantially improves the model's performance, while for Ethereum and Tether, it leads to a slight decrease in precision. In particular, using only Group No.~1 is sufficient for predicting Ethereum and Tether, as this group has the highest performance rate. For Bitcoin, introducing technical indicators slightly improves the performance. It is worth noting that DBN~(OHLCV-TI) consistently outperforms or matches those incorporating all features. 

The impact of external factors on the prediction performance of DBN models has mixed results. While introducing external factors~(OHLCV-EF), which contains tweet numbers and traditional financial assets, to the DBN model constructed by OHLCV improves the performances for predicting Binance Coin and Litecoin, it reduces the precision of the model outputs for Bitcoin, Ethereum and Ripple.
In the case of Tether, the performance of both DBN~(OHLCV) and DBN~(OHLCV-EF) remains unchanged. Considering its DBN structure, these external factors do not influence the close price as they are isolated in the DBN model, and no arc connects them to OHLCV. Additionally, it should be noted that incorporating external factors along with OHLCV and technical indicators to construct all features does not consistently result in improved model precision. For example, in the case of Ripple, the DBN model constructed with all features~(OHLCV-EF-TI) has lower performance than the DBN model constructed by combining OHLCV and technical indicators~(OHLCV-TI). This indicates that not all external factors may have a significant impact on the prediction accuracy of the model. Hence, selecting the most relevant features is crucial for improving the precision of the DBN models. The results presented are visualised Figure~\ref{fig:percision-DBNs} for a better understanding of the results. One important observation is that Tether shows the lowest precision across all the cryptocurrencies. Moreover, the highest precision is observed for the DBN~(OHLCV-TI) model applied to Ripple, while the lowest precision is recorded for the DBN~(OHLCV-EF-TI) model in the case of Tether. 



\begin{figure*}[h]
  \centering
\includegraphics[width=\linewidth,height=\textheight,keepaspectratio]{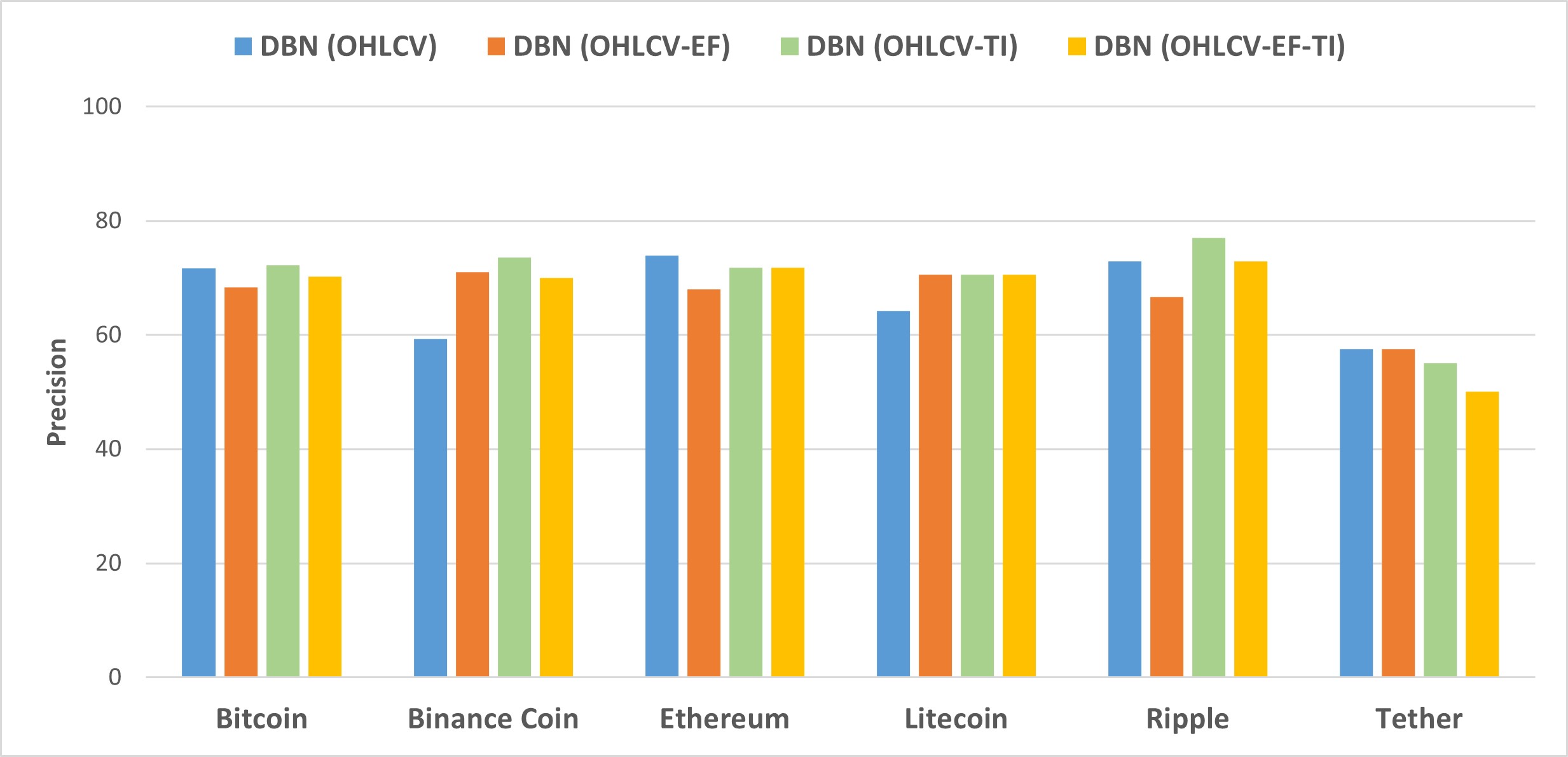}
 \caption{Precision in prediction based on different feature groups evaluates the performance of constructed DBNs}
\label{fig:percision-DBNs}
\end{figure*}

\subsection{Analysing causal structures}

Analysing the structures of DBNs in Figure~\ref{fig:DBN} offers essential insights, enabling investors to improve predictions by identifying key factors affecting cryptocurrency prices, and demonstrating their advantage in explainability over other AI deep learning methods. 

\begin{figure*}[h]
 \centering
\begin{subfigure}{.5\textwidth}
  \centering
\fbox{\includegraphics[ width=0.96\linewidth,height=\textheight,keepaspectratio]
 {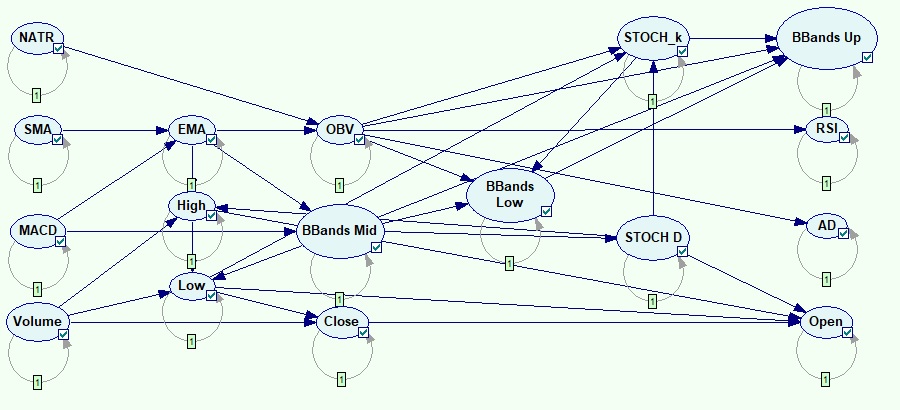}}
  \caption{Bitcoin DBN~(OHLCV-TI)}
  \label{fig:BTC_DBN}
\end{subfigure}%
\begin{subfigure}{.5\textwidth}
  \centering
\fbox{\includegraphics[ width=0.96\linewidth,height=\textheight,keepaspectratio]
 {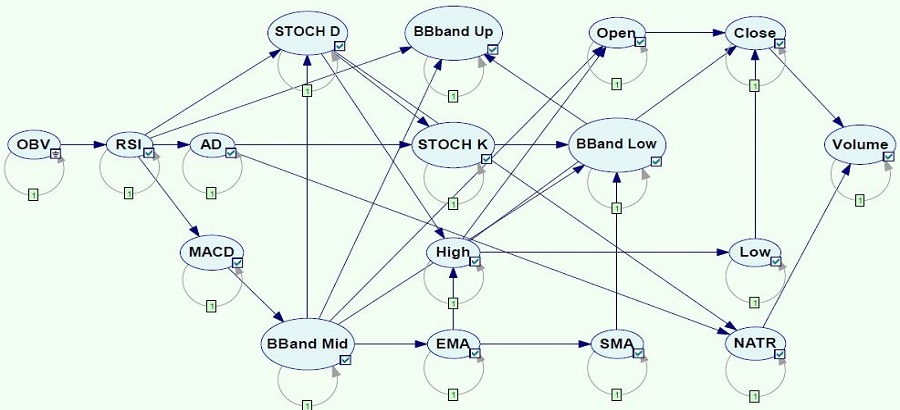}}
  \caption{Binance Coin DBN~(OHLCV-TI)}
  \label{fig:BNB_DBN}
\end{subfigure}
\begin{subfigure}{.5\textwidth}
   \centering
\fbox{\includegraphics[width=0.96\linewidth,height=\textheight,keepaspectratio]{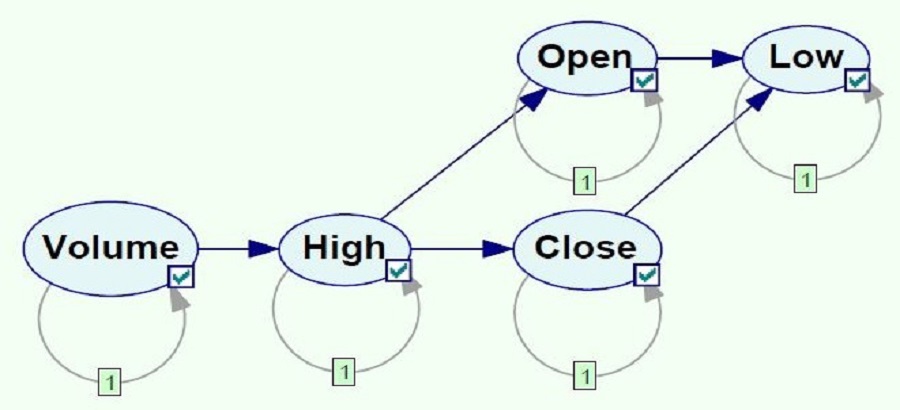}}
  \caption{Ethereum DBN~(OHLCV)}
  \label{fig:ETH_DBN}
\end{subfigure}%
\begin{subfigure}{.5\textwidth}
   \centering
\fbox{\includegraphics[width=0.96\linewidth,height=\textheight,keepaspectratio]{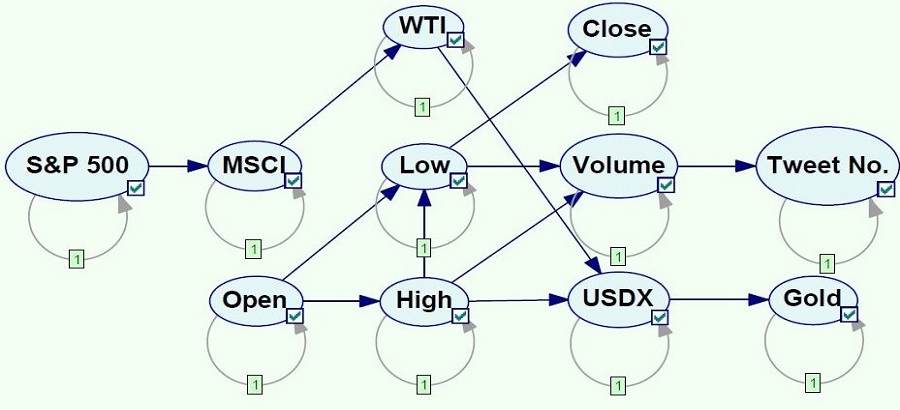}}
  \caption{Litecoin DBN~(OHLCV-EF)}
  \label{fig:LTC_DBN}
\end{subfigure}
\begin{subfigure}{.5\textwidth}
  \centering
\fbox{\includegraphics[width=0.96\linewidth,height=\textheight,keepaspectratio]{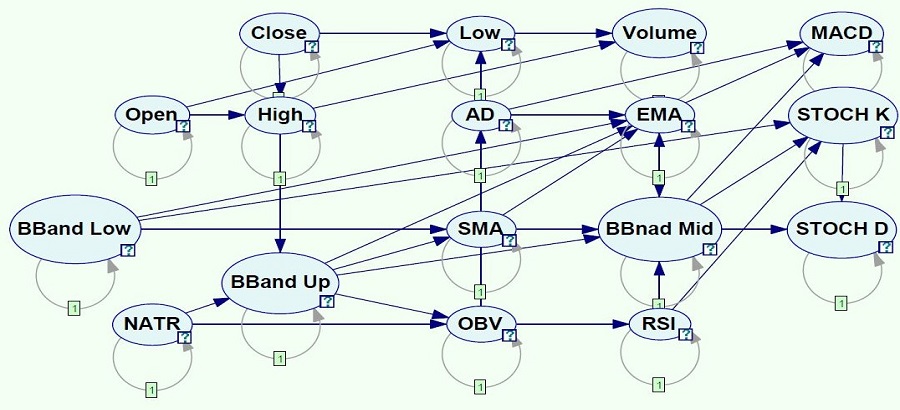}}
  \caption{Ripple DBN~(OHLCV-TI)}
  \label{fig:XRP_DBN}
\end{subfigure}%
\begin{subfigure}{.5\textwidth}
  \centering
\fbox{\includegraphics[width=0.96\linewidth,height=\textheight,keepaspectratio]{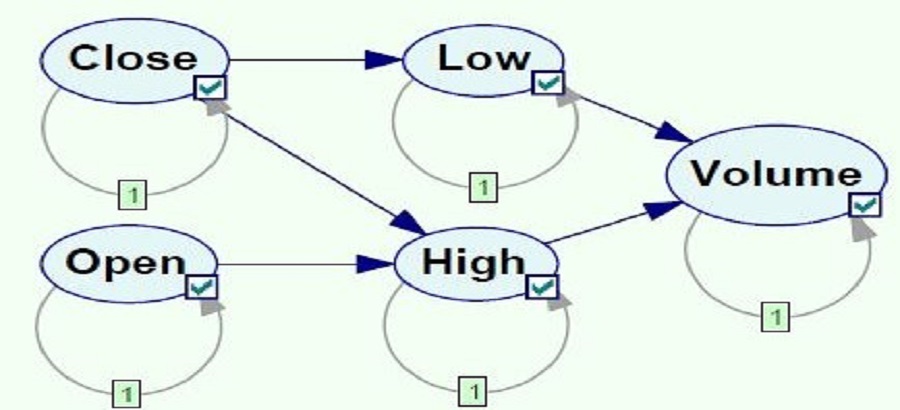}}
  \caption{Tether DBN~(OHLCV)}
  \label{fig:Tether_DBN}
\end{subfigure}
 \caption{The DBNs with the highest precision for each altcoin, considering a temporal plate of five-time steps. The numbers inside the parentheses indicate the most influential set of feature categories as represented in Table \ref{tab:preds-DBNs}} 
\label{fig:DBN}
\end{figure*}

The figure displays the best-performing DBNs associated with each cryptocurrency in Table~\ref{tab:preds-DBNs}.\footnote{Note, the DBNs are chosen using the precision metric.} 
Bitcoin, Binance Coin, and Ripple achieve best performance with DBN~(OHLCV-TI) (Sub-figures~\ref{fig:BTC_DBN},~\ref{fig:BNB_DBN} and \ref{fig:XRP_DBN}), while Ethereum and Tether perform best with DBN~(OHLCV) (Sub-figures~\ref{fig:ETH_DBN} and \ref{fig:Tether_DBN}). Litecoin, on the other hand, stands out as the only coin with its best performance in DBN~(OHLCV-EF)(Sub-figure~\ref{fig:LTC_DBN}).
Notably, all nodes within each DBN are interconnected, without any isolated nodes. This highlights the inherent interdependencies among these entities and their mutual influence on each other, specifically for the selected combination of feature categories. Moreover, the structure of DBNs varies across different cryptocurrencies, even within the same group of features, underscoring the unique characteristics of each coin. For instance, when examining Sub-figures~\ref{fig:ETH_DBN} and \ref{fig:Tether_DBN}, the causal relationships between the components of OHLCV features for Ethereum and Litecoin exhibit different dynamics. This distinction further emphasises the individual nature of each cryptocurrency and the specific factors that influence their price movements.

Furthermore, in Sub-figure~\ref{fig:BNB_DBN}, the DBN representing Binance Coin, the root node OBV exhibits the highest influence, evidenced by its numerous descendants. Additionally, RSI and BBand mid nodes display significant interactivity, possessing the highest number of edges connecting them to other nodes. Contrastingly, Sub-figure~\ref{fig:ETH_DBN}, which showcases the best-performing DBN for Ethereum, reveals the Volume node as the root node with the most descendants. Among the other nodes, High exhibits the highest level of interactivity. Furthermore, regardless of the set of feature categories employed, the best-performing DBNs for Ethereum~(Sub-figure~\ref{fig:ETH_DBN}) and Tether~(Sub-figure~\ref{fig:Tether_DBN}) exhibit distinctly different structures in terms of node interactions. This emphasises the unique relationships between the nodes within each DBN and underscores the influence of specific factors on the price dynamics of Ethereum and Tether. The same argument is valid for Bitcoin~(Sub-figure~\ref{fig:BTC_DBN}), Binance coin~(Sub-figure~\ref{fig:BNB_DBN}) and Ripple~(Sub-figure~\ref{fig:XRP_DBN}) as they share the same category of features with completely different dynamics among their respective nodes. 



\section{Conclusions and future work}\label{Conclusions}

This study assesses the effectiveness of DBNs in predicting the price movements of Bitcoin and five major altcoins. DBNs consistently outperform ARIMA, SVR, LSTM, RF, and SVM models, showing an average precision improvement of at least 15\%, however, performance varies by cryptocurrency due to differing market dynamics.

The second goal of this study is to examine how feature selection affects model performance by using various combinations of four groups of 23 features. The results show that while combining basic price information with technical indicators yields the most accurate predictions, using only basic price information results in the lowest prediction scores. Additionally, increasing the number of features does not always lead to better precision, highlighting the need for careful feature selection to optimise DBN performance in cryptocurrency price prediction.

The findings of this study can contribute to developing a reliable decision-support system, enhancing investment accuracy and optimising trading strategies. Future research could explore integrating expert elicitation with DBNs, combining expert insights with data-driven methods to reduce subjectivity and reliance on large datasets. Additionally, incorporating internal factors such as blockchain data alongside external market influences could further improve prediction accuracy.

One important factor impacting AI prediction models is the frequency of timeframes used for features. While this study uses intraday price data, exploring higher frequencies, like four-hour or one-hour intervals, could affect DBN precision. Future research could investigate how various timeframes influence DBN performance under different market conditions, including bull and bear markets, where cryptocurrencies behave differently.

\appendix
\section{A typical scenario for DBNs}\label{sampleDBN}

In this appendix, we conduct additional analysis to gain a deeper understanding of the behaviour of DBNs. The networks for Ethereum and Tether
 illustrated in Sub-figures~\ref{fig:ETH_DBN} and \ref{fig:Tether_DBN}, respectively, have the smallest number of nodes compared to the other networks. This characteristic makes them suitable for further visual examination of the interactions between their nodes. Therefore, we showcase a typical scenario of the DBNs for Ethereum and Tether, where we fix the states of the nodes, excluding the close node, and observe the resulting changes in probabilities for the target node~(close price). Table~\ref{scenario}, Sub-figures \ref{fig:ETH_DBN_state} and \ref{fig:Tether_DBN_state} illustrate the changes in the close price.

\begin{table*}[h]
\caption{\textbf{Typical scenario}~-~ Five-day movement directions for Ethereum and Tether considering open, high, and low prices along with changes in the respective volume. The ``Prob.''column displays the probabilities of the closing price direction being up or down on the fifth day, derived from the observed scenarios}
\label{scenario}
\renewcommand{\arraystretch}{1.2}
 \resizebox{\columnwidth}{!}{%
\begin{tabular}{|c|l|lllll|r|c|}
\hline
\textbf{Cryptocurrency} & \textbf{Feature} & \multicolumn{5}{c|}{\textbf{Five day movement scenario}} & \textbf{Prob. of Up Close}& {\textbf{Prob. of Down Close}} \\ \hline  
\multirow{4}{*}{Ethereum} & Open & \multicolumn{1}{l|}{Up} & \multicolumn{1}{l|}{Up} & \multicolumn{1}{l|}{Down} & \multicolumn{1}{l|}{Down} & Down & \multicolumn{1}{c|}{\multirow{4}{*}{0.16241266}} & \multirow{4}{*}{0.83758734} \\ 
& High & \multicolumn{1}{l|}{Down} & \multicolumn{1}{l|}{Up} & \multicolumn{1}{l|}{Down} & \multicolumn{1}{l|}{Down} & Down & \multicolumn{1}{l|}{} &  \\ 
 & Low & \multicolumn{1}{l|}{Up} & \multicolumn{1}{l|}{Down} & \multicolumn{1}{l|}{Down} & \multicolumn{1}{l|}{Down} & Down & \multicolumn{1}{l|}{} &  \\ 
 & Volume & \multicolumn{1}{l|}{Down} & \multicolumn{1}{l|}{Up} & \multicolumn{1}{l|}{Down} & \multicolumn{1}{l|}{Up} & Up & \multicolumn{1}{l|}{} &  \\ \hline
\multirow{4}{*}{Tether} & Open & \multicolumn{1}{l|}{Up} & \multicolumn{1}{l|}{Up} & \multicolumn{1}{l|}{Up} & \multicolumn{1}{l|}{Down} & Down & \multicolumn{1}{c|}{\multirow{4}{*}{0.49966089}} & \multirow{4}{*}{0.50033911} \\ 
 & High & \multicolumn{1}{l|}{Down} & \multicolumn{1}{l|}{Up} & \multicolumn{1}{l|}{Up} & \multicolumn{1}{l|}{Down} & Up & \multicolumn{1}{l|}{} &  \\ \
 & Low & \multicolumn{1}{l|}{Up} & \multicolumn{1}{l|}{Down} & \multicolumn{1}{l|}{Up} & \multicolumn{1}{l|}{Down} & Down & \multicolumn{1}{l|}{} &  \\ 
 & Volume & \multicolumn{1}{l|}{Up} & \multicolumn{1}{l|}{Up} & \multicolumn{1}{l|}{Down} & \multicolumn{1}{l|}{Down} & Up & \multicolumn{1}{l|}{} &  \\ \hline
\end{tabular}
}
\end{table*}

As it can be seen in Figure~\ref{fig:DBN_state}, in both cryptocurrencies, the probabilities of the close nodes changing to the down state, shown in purple, dominate. Therefore, the predictions of the DBNs indicate a downward movement for both coins. Notably, the probability of the close state being down is higher for Ethereum compared to Tether.

\begin{figure*}[h]
 \centering
\begin{subfigure}{.5\textwidth}
  \centering
\fbox{\includegraphics[ width=\linewidth,height=\textheight,keepaspectratio]
 {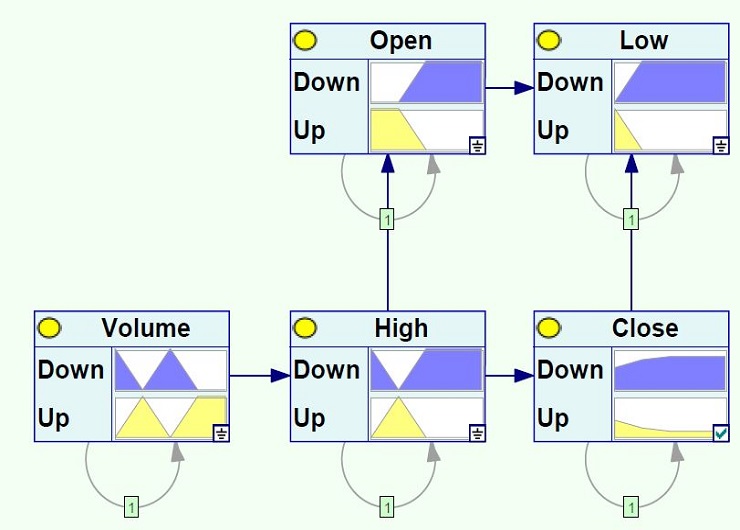}}
  \caption{Ethereum DBN~(OHLCV)}
  \label{fig:ETH_DBN_state}
\end{subfigure}%
\begin{subfigure}{.5\textwidth}
  \centering
\fbox{\includegraphics[ width=0.98\linewidth,height=0.98\textheight,keepaspectratio]{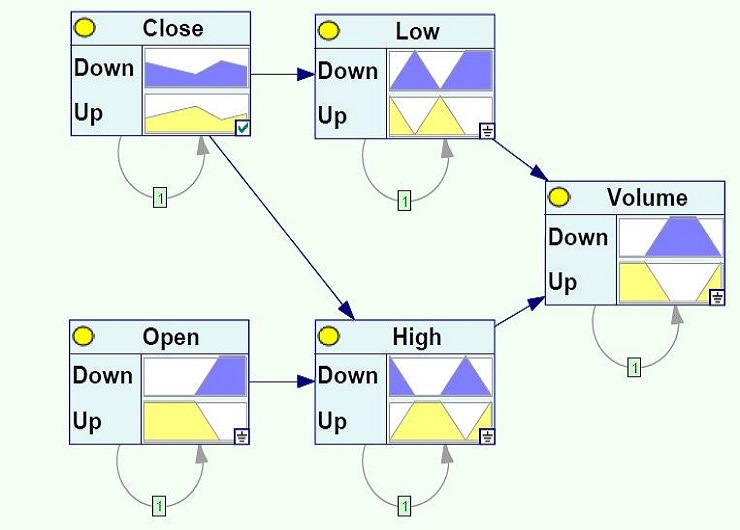}}
  \caption{Tether DBN~(OHLCV)}
  \label{fig:Tether_DBN_state}
\end{subfigure}
 \caption{Typical scenario analysis for predicting price directions in the best-performing DBNs for Ethereum and Tether} 
\label{fig:DBN_state}
\end{figure*}

\section{Data}
The data of this study are available in the Git repository at:\\ \url{https://github.com/bigrasam/Dynamic-Bayesian-Networks-Paper-Data.git}

\section{Descriptive analysis of data frames}\label{Descriptive}
This section provides a detailed descriptive analysis of the closing price data and daily tweet volumes for cryptocurrencies and traditional financial assets, highlighting key statistics and trends. Visualisations further illustrate the variations and patterns in price fluctuations and sentiment over time.

Descriptive statistics of close price data of cryptocurrencies and traditional financial assets are presented in Table~\ref{tab:data description}. As the table shows, Bitcoin exhibits the highest mean close price among cryptocurrencies, indicating its relatively higher value than others. Meanwhile, Binance Coin's higher standard deviation suggests that its price fluctuates more widely over time, indicating greater price volatility. 
\begin{table*}[htbp]
 \centering
 \caption{Summary statistics for the dataset of combined daily close prices of cryptocurrencies and traditional financial assets. The data period is between January 2018 and October 2022. The last column ``Obs.'' specifies the total number of observations available at the time of extracting data}  
 \label{tab:data description}
 \resizebox{\columnwidth}{!}{%
\renewcommand{\arraystretch}{1.1}
 \begin{tabular}{lllllllr}
 \hline
  \textbf{} & \textbf{Mean} & \textbf{Std. Dev.} & \textbf{Min.} &\textbf{Median}&\textbf{Max.} & \textbf{Obs.}\\ \hline
  \textbf{Bitcoin} & 20705.53 &17356.86 &3242.48 &10796.95& 67566.83 &1189\\ 
   \textbf{Binance coin} & 151.56 &187.21 & 4.53 &27.22&675.69 &1171\\ 
  \textbf{Ethereum} &1150.13  &1225.25  &884.31  & 474.21 &4812.09&1206\\ 
  \textbf{Litecoin}  &101.55  &64.61 &75.172 &23.47&377.39 & 1206 \\  
  \textbf{Ripple}  &0.52 	 &0.33  &0.14&0.38&2.456 &1184 \\
   \textbf{Tether}  &1.01 	 &0.01  &0.97 	 &1.02 &1.09&1232\\
   \hline
    \textbf{Gold} & 16.79 & 4.5 & 8.8 &16.09 &28.17&1464\\ 
  \textbf{MSCI } & 2360.99 & 410.09& 1596 & 2197.15 &3242.3& 1464\\ 
  \textbf{S\&P 500}  & 3265.73 & 718.52 &2237.4 &2985.61 &4796.56&1464 \\
   \textbf{USDX}  &96.36 & 4.61&88.59 & 96.13&114.11&1464 \\
   \textbf{WTI}  &61.91 & 19.02 & -37.63&58.82 & 123.7&1464\\ 
   \hline
    \end{tabular}
}
\end{table*}

Table~\ref{tab:tweet_describtion} provides summary statistics for the daily tweet numbers associated with each cryptocurrency in this study between January 2018 and October 2022. As previously mentioned, daily tweet data associated with Tether is not available on www.bitinfocharts.com, and hence Tether is not presented in Table~\ref{tab:tweet_describtion}.

\begin{table*}[htbp]
 \centering
 \caption{Summary statistics for daily tweet number associated with the cryptocurrencies in this study between January 2018 and October 2022}
 \label{tab:tweet_describtion}
 \resizebox{\columnwidth}{!}{%
\renewcommand{\arraystretch}{1.1}
 \begin{tabular}{llllllll}
 \hline
  \textbf{} & \textbf{Mean} & \textbf{Std. Dev.} & \textbf{Min.} &\textbf{Median}&\textbf{Max.}\\ \hline
  \textbf{Bitcoin} &55551.35  &48305.20  &445  &31428.50&363566\\
    \textbf{Binance coin} & 120.05 &  215.83 & 1 &55 &1601\\ 
  \textbf{Ethereum} & 17912.37 & 15881.35 & 2418 & 12058 &138220 \\ 
  \textbf{Litecoin}  & 1544.24 & 1301.59 &360 & 1172& 13778 \\  
  \textbf{Ripple}  & 	17233.21& 39085.54 & 	2362 &7498 &735252 \\ \hline
\end{tabular}
}
\end{table*}

Figure~\ref{fig:plots_price} shows the close prices of cryptocurrencies and traditional financial assets from January 2018 to October 2023. We see that all traditional financial assets experienced growth during this period. However, the close price plots of Binance Coin and Ethereum demonstrate a relatively steady trend before 2021, while Litecoin and Ripple experienced a considerably unstable period during 2018 before a period of stability until 2021.
In 2021, all coins experienced high variation in their close prices, except  Tether, as it serves as a stablecoin. For instance, Ethereum's close price variances were 44931.208 and 920876.95 before 2021 and after 2021, respectively. As a result, a high level of prediction error can be expected due to the increasing variances in data. 

\begin{figure*}[htbp]
\centering
\includegraphics[width=1.02\linewidth,height=\textheight,keepaspectratio]{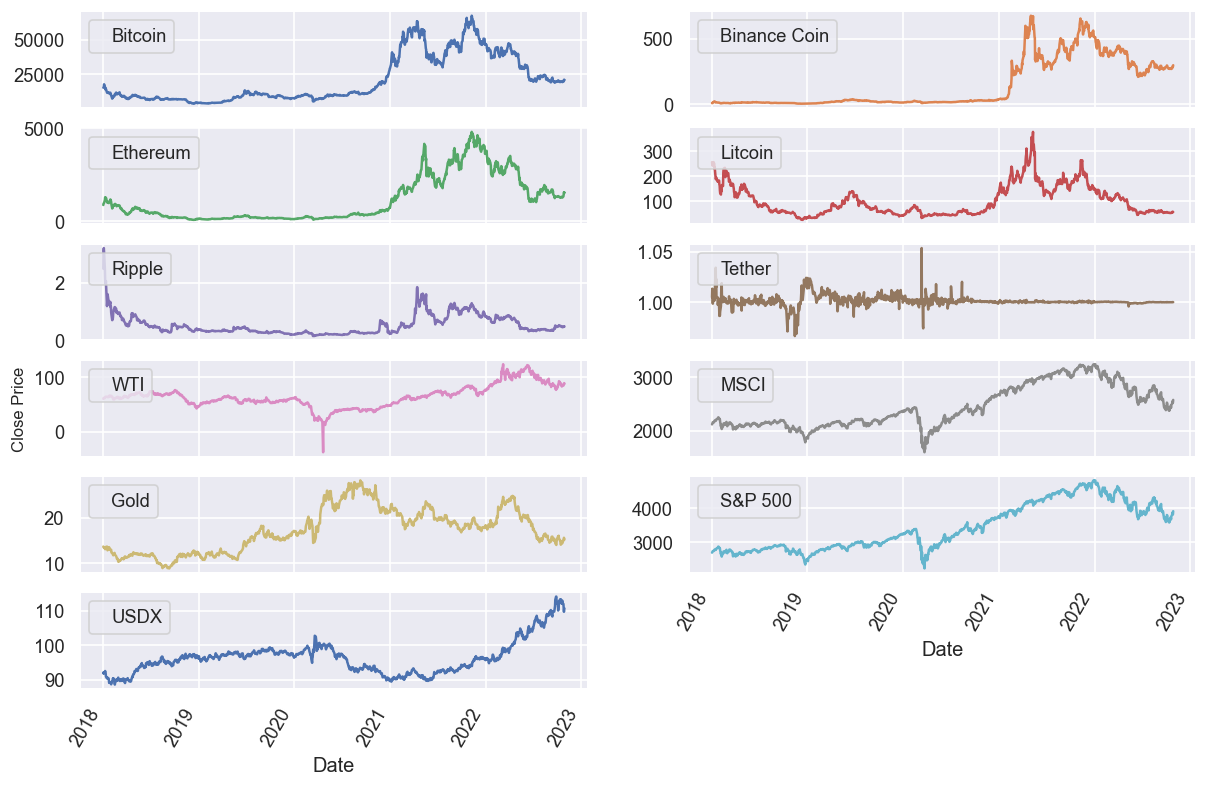}
\caption{The price fluctuation of traditional financial assets and cryptocurrencies in this study}
\label{fig:plots_price}
\end{figure*}

Figure~\ref{fig:tweet_no} shows the daily tweet numbers for all the coins. Despite several spikes, the trends of tweet numbers are relatively consistent compared to the price data in Figure~\ref{fig:plots_price}. We observe a significant jump in price during 2021 compared to other years

\begin{figure*}[h]
  \centering
\includegraphics[width=\linewidth,height=\textheight,keepaspectratio]{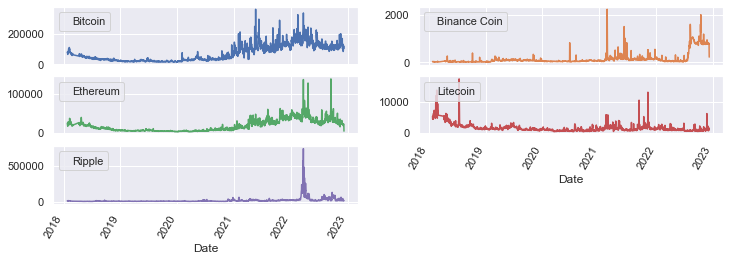}
  \caption{The number of daily tweets associated with Bitcoin, Binance Coin, Ethereum, Litecoin, and Ripple between January 2018 and October 2022}
  \label{fig:tweet_no}
\end{figure*}

 \section{Description of selected features}\label{feature_definitions}
This section provides details of the features used in the study for predicting cryptocurrency price movements.

\begingroup
\renewcommand{\arraystretch}{1.1} 
\begin{table}[]
\footnotesize
 \centering
\caption{
Description of selected  features, their types, and the time window considered in this study}
\label{tab:features}
\begin{tabular*}{\textwidth}{|p{6pc}|p{3.5pc}p{13.9pc}>{\raggedleft\arraybackslash}p{5.1pc}|}
\hline
\multicolumn{1}{|c|}{\textbf{Type}} &
  \multicolumn{1}{c}{\textbf{Feature}} &
  \multicolumn{1}{c}{\textbf{Description}} &
  \textbf{Time window} \\ \hline

\multirow{8}{*}{Macro-financial}& Gold&  The gold spot market price in US dollars  &
  Daily   \\
& MSCI  & A market capitalisation-weighted index comprising 1,546 companies from around the world &
  Daily            \\
& S\&P500 & A market capitalisation-weighted index of the 500 leading publicly traded companies in the US &
  Daily\\
& USDX &
  The value of the US dollar relative to a basket of six foreign currencies &
  Daily            \\
& WTI  &
   A popular oil price benchmark &
  Daily              \\ \hline
\multirow{8}{*}{OHLVC}   &  Close price &
  The price at which a cryptocurrency  is last \newline traded in a trading interval   &  Daily        \\
& Low price  &
  It is a cryptocurrency's lowest trading price  in a trading interval &
 Daily\\
& High price &  A cryptocurrency's highest trading price  in a trading interval &
  Daily\\
& Open price   &
  The price at which a cryptocurrency is first traded in a trading interval &
  Daily\\
& Volume &   The total number of cryptocurrencies traded  in a trading interval &
  Daily\\ \hline
 \multirow{2}{*}{Social media} & Tweet number & The number of daily tweets associated with a cryptocurrency& Daily\\ \hline
\multirow{18}{*}{\shortstack{Technical\\indicators}} & AD & It uses volume and price to calculate the money flow into or out of a security and determines the accumulation or distribution of funds by traders &
  Last Period            \\
& BBands &  It consists of a band of three lines, usually SMA in the middle, and the upper and lower bands are positioned two standard deviations away from the SMA &
  5 days            \\
& EMA &It uses moving averages; however, it applies more weight to recent data points to reduce the data lag &
   10 days           \\
& MACD &
MACD measures two EMAs  (typically EMA for 12 and 26 days) &
     Fast= 12, slow= 26, signal= 9 \\
& NATR    &
  It is a metric to measure volatility. &
  14 day          \\
& OBV  &
  OBV is a cumulative indicator that measures the buying and selling pressure &
  Last period             \\
& RSI &It indicates overbought and oversold conditions by comparing the magnitude of gains and losses in stocks &
  14 days             \\
& SMA&  The SMA average data points for a given period. & 10 days           \\
& Stoch   &
  It is a range-bound momentum indicator that potentially determines overbought and oversold situations. &
  14 days   \\  \hline  
\end{tabular*}
\end{table}
\endgroup

\end{document}